\documentclass{article}
\usepackage{arxiv}
\usepackage[utf8]{inputenc} 
\usepackage[T1]{fontenc}    
\usepackage{hyperref}       
\usepackage{url}            
\usepackage{booktabs}       
\usepackage{amsfonts}       
\usepackage{nicefrac}       
\usepackage{microtype}      
\usepackage{lipsum}		
\usepackage{graphicx}
\usepackage{doi}
\usepackage{array}
\usepackage{setspace}
\graphicspath{{images/}}
\usepackage{siunitx}
\usepackage{tabularx}
\usepackage{authblk}
\usepackage{makecell}
\usepackage{chemformula}

\title{Anomaly segmentation model for defects detection in electroluminescence images of heterojunction solar cells}

\date{May 26, 2022}	

\author[1, *]{Alexey Korovin}
\author[1]{Artem Vasilyev}
\author[2]{Fedor Egorov}
\author[2]{Dmitry Saykin}
\author[3]{Evgeny Terukov}
\author[4]{Igor Shakhray}
\author[5]{Leonid~Zhukov}
\author[1, 6]{Semen~Budennyy}

\affil[1]{Artificial Intelligence Research Institute (AIRI)}
\affil[2]{Hevel LLC, Russia}
\affil[3]{Saint Petersburg Electrotechnical University "LETI"}
\affil[4]{Unigreen Energy LLC}
\affil[5]{HSE University}
\affil[6]{Sber AI Lab, Moscow}
\affil[*]{Corresponding author: Alexey Korovin, korovin@airi.net}

\renewcommand{\shorttitle}{\small\raggedright{Anomaly segmentation model for defects detection in electroluminescence images of heterojunction solar cells}}

\begin{document}
\maketitle

\begin{abstract}
Efficient defect detection in solar cell manufacturing is crucial for stable green energy technology manufacturing. This paper presents a deep-learning-based automatic detection model SeMaCNN for classification and semantic segmentation of electroluminescent images for solar cell quality evaluation and anomalies detection. The core of the model is an anomaly detection algorithm based on Mahalanobis distance that can be trained in a semi-supervised manner on imbalanced data with small number of digital electroluminescence images with relevant defects. This is particularly valuable for prompt model integration into the industrial landscape. The model has been trained with the on-plant collected dataset consisting of 68 748 electroluminescent images of heterojunction solar cells with a busbar grid. Our model achieves the accuracy of 92.5\%, F1 score 95.8\%, recall 94.8\%, and precision 96.9\% within the validation subset consisting of 1049 manually annotated images. The model was also tested on the open ELPV dataset and demonstrates stable performance with accuracy 94.6\% and F1 score 91.1\%. The SeMaCNN model demonstrates a good balance between its performance and computational costs, which make it applicable for integrating into quality control systems of solar cell manufacturing.
\end{abstract}

\keywords{solar cells \and image processing \and deep learning \and computer vision \and diagnostics}

\section{Introduction}
Solar energy plays a crucial role in the transition towards a renewable and carbon-neutral power supplies for sustainable development by ESG principles (Environmental, Social and Corporate Governance)\cite{noauthor_esg_nodate}. This proven technology demonstrates an exponential pace of deployment and is expected to supply petawatts of energy within the following years. One of the key factors driving this trend is a continuous price decline of photovoltaic panels together with the steady increase of their cumulative installed capacity over the last decade\cite{adrian_whiteman_renewable_2020}.
A photovoltaic cell (PV) is a basic unit for converting solar energy into electricity. A set of solar cells are assembled and interconnected into a solar panel to provide electric power for commercial applications.\cite{tsai_defect_2013}
Solar cells manufacturing, though mature, still experiences process faults. High technological process variability during manufacturing like chemical treatment, doping, deposition of transparent conducting and metal coatings, mechanical forces - all can damage PV. This in turn may eventually shorten the lifespan of the solar cell due to thermal deterioration or losing its performance characteristics: short circuit current (ISC), open-circuit voltage (VOC) and fill-factor (FF) etc. Efficient defects monitoring of solar cells during the manufacturing process can prevent defective cells from being used in subsequent panel assembly, help to manage solar cells deterioration, enhance their performance, improve reliability and durability.\cite{tsai_defect_2013}
Electroluminescence (EL) diagnostics, along with photoluminescence and lock-in thermography, are the widespread techniques for visual detecting and mapping of solar cell defects within quality control systems\cite{breitenstein_quantitative_2010}. Electroluminescence utilizes infrared range cameras to capture the cell or panel images. In the EL imaging system, voltage applied to a solar module in dark environment induces current and emission of infrared light then captured by a cooled Si-CCD or InGaAs camera. High peak current required for high-resolution electroluminescence images acquisition may cause degradation of solar cells. Intensity of emission is higher in regions with high conversion efficiency and lower at defective sites like intrinsic crystal grain boundaries and extrinsic defects, including micro-cracks, breaks, and finger interruptions because defects are inactive and hardly emit light\cite{tsai_defect_2013}. Images with mapped intensity of electroluminescent light enable engineers to analyze, troubleshoot or predict faults using defect features. 
Manual inspection of EL images is labor-consuming, the inspection accuracy may fluctuate due to the tedious and routine nature of the process and is not possible for high throughput manufacturing lines. Consequently, these conditions make advanced statistical models more attractive to enhance the PV diagnostics process in high throughput PV manufacturing lines. These models are mainly divided into conventional machine learning and deep learning models. 
Conventional techniques like independent component analysis are able to detect the presence of cracks, scratches, and finger interruption in PV modules, but not differentiate them\cite{tsai_defect_2013}. Other available approaches include anisotropic diffusion filtering,\cite{anwar_micro-crack_2014} matched filtering,\cite{spataru_automatic_2016} and vesselness filtering\cite{stromer_enhanced_2019} for successful micro-crack classification in EL images. 
Since defects of different types have significantly varying patterns, single image processing approach can not successfully handle all of them. At the same time, deep learning models have become more appropriate for modern diagnostics systems capable to detect a great number of defect groups.\cite{ahmad_photovoltaic_2020,bartler_automated_2018,mansouri_deep_2021,zhao_deep_2021,deitsch_automatic_2019,mayr_weakly_2020,akram_cnn_2019,lin_efficient_2021} 
Deep learning models based on convolutional neural networks (CNN) often applied to computer vision tasks and outperform traditional feature extraction based models like support vector machines.\cite{ahmad_photovoltaic_2020,mansouri_deep_2021,deitsch_automatic_2019} For example, the resNet50 model is able to segment cracks in images using weakly-supervised learning on a limited dataset\cite{mayr_weakly_2020} and perform classification of up to five defect types\cite{akram_cnn_2019}. CNNs capable of multi-type classification are also reported\cite{zhao_deep_2021}, however, the accuracy of proper classification of top-14 possible defect types was 70.2\% $mAP_{50}$ ($mAP_{50}$ mean average precision with at least 50\% localization accuracy). 
Deep learning models benefit from automated generation of features. Feature space is generated by application of specific operations (convolutional, pooling, etc.) to images to provide highest correlation with the desired parameter, presence and type of defects. Deep models also provide the possibility of transfer learning: once the model is trained on the dataset of one manufacturing line, its application to another line is possible with minimal additional training. Advantages of models adopting deep learning architecture can be tremendous when sufficient data is available even without deploying pre-designed algorithms for feature extractions. 
Training of CNNs designer for computer vision tasks is challenging. EL image datasets are highly imbalanced due to a relatively small fraction of defective cells. Furthermore, the minority class shows significant variations in the structure and position of the defects.\cite{bartler_automated_2018} To improve the performance of diagnostics models, image pre-processing approaches like cell extraction, distortion correction, perspective correction, and image augmentation by mirroring and flips are used.\cite{bartler_automated_2018,lin_efficient_2021} Hence, on the one hand, classical machine learning models, widely used for quality control, require complex features adjustment or engineering, on the other hand, the use of supervised deep learning algorithms involves labor-intensive data annotation. In this paper, we present a deep learning detection model SeMaCNN that performs both semantic segmentation i.e., highlighting defect regions of all types and classification, i.e. cells quality assessment. 
SeMaCNN is a semi-supervised model combined anomaly detection algorithms via unsupervised feature extractor and supervised shallow classifier. This is particularly useful in ordinary manufacturing conditions when a large dataset is available with only a small labeled part. Unsupervised feature extractor is implemented via PaDiM\cite{defard_padim_2020} model with Mahalanobi distances for semi-orthogonal embedding and shallow classifier via ResNet18 convolution neural network. 
The model was trained on the dataset consisting of 68 748 on-plant collected electroluminescent images of heterojunction solar cells with 1049 manually annotated samples, and achieved an accuracy of 92.5\%, F1 score of 95.8\%, recall of 94.8\%, and precision of 96.9\%. The model also demonstrated a solid performance on the open ELPV dataset\cite{buerhop-lutz_benchmark_2018} with accuracy of 94.6\% and F1 score of 91.1\%.

\section{Description of datasets and defect detection models}
\subsection{Dataset description}
\subsubsection{Hevel Electroluminescence Photovoltaic Images Dataset (HEPV)}
\begin{figure}[ht]
\center{\includegraphics[scale=1]{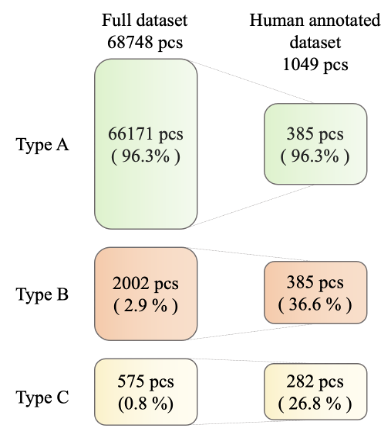}}
\caption{Composition of HEPV dataset of solar cell EL images. The whole dataset consisted of 66171 images of class A, 2002 of class B, and 575 images of class C (499 images with dark spots and 76 images with cracks). A subset of 1049 randomly selected images (385 of class A, 385 of class B, 206 of class C with dark spots, 76 of class C with cracks).}
\label{im: figure1}
\end{figure}

\begin{figure}[ht]
\center{\includegraphics[scale=1]{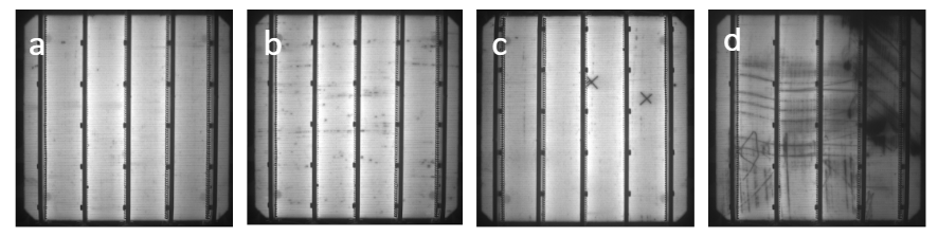}}
\caption{Examples of EL images of BB cells:  a) defect-free cell of quality class A; b) cell with slight deterioration spots not still considered to reduce operating power and lifespan significantly of B-class quality; c) Cell contains a cross-shaped crack of C class quality, d) Cell contains a combination of dark deterioration areas and scratches of C class quality.}
\label{im: figure2}
\end{figure}

In this work, we present the classification method for PV heterojunction batteries with busbar (BB) grid type. PV cells have been divided by manufacturer into three groups based on a combination of EL characteristics corresponding to maximum power and expected lifespan of a panel: A – fully operational PV cell, B – a subprime cell (still can be used for panel assembly (consider 0.75\% defect probability) and C – defective cell (rejected for further utilization).
The dataset consists of 68~748 EL images of solar cells of type BB collected on PV-IUCT-3600 (Halm) with Cetis PV-EL package at 3V, 12A with about 17 ms exposition time. All images in the dataset have been labeled by the manufacturer’s proprietary deterministic system with recorded accuracy of 96.1\%. A subset of 1049 randomly selected images has been manually annotated for validation purposes and re-balanced. The dataset composition is depicted in \ref{im: figure1}.
The goal of the study is to classify solar cells into production classes and to detect various types of defects according to the deterministic (threshold-based) manufacturing system. Only less than 10\% of images in the dataset are fully defect-free (\ref{im: figure2}a) and classified as entirely acceptable (class A) and more than 90\% of images contain some defects. Defects such as small gray and black dots (\ref{im: figure2}b), backside contamination and chemical defects are classified into class A and B. Cells with substantial dark spots, as well as any type of scratches and cracks (\ref{im: figure2}c) or with more pronounced manifestations of minor defects (\ref{im: figure2}d) considered unacceptable.

\subsubsection{ELPV Dataset}
\begin{figure}[ht]
\center{\includegraphics[scale=1]{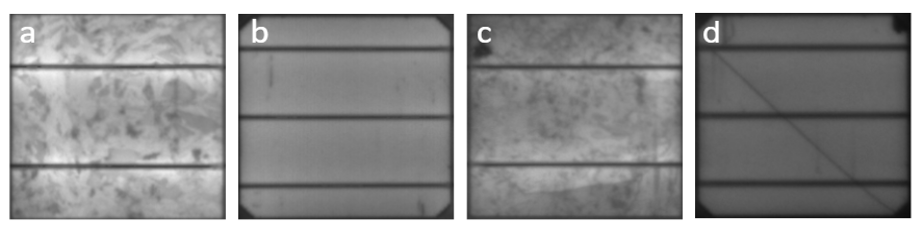}}
\caption{Samples of EL images of solar cells from ELPV dataset: a) defect certainty = 0, b) defect certainty = 0.33, c) defect certainty = 0.66, d) defect certainty = 1.}
\label{im: figure3}
\end{figure}

The ELPV dataset is an open dataset for the anomaly detection and classification of photovoltaic cells. This dataset was presented in Buerhop-Lutz et al.\cite{buerhop-lutz_benchmark_2018} and suggested as a benchmark. The dataset contains 1116 images of working solar cells and 1508 images of defective solar cells. In addition to the binary classification of images, this dataset estimates the confidence level of that classification for each image. 
To better evaluate model performance and compare model performance across datasets we binarized confidence level values with a zero threshold for general task performance and with a threshold at 0.33 for comparing performance across datasets.\cite{otamendi_segmentation_2021} Examples of EL images with corresponding defect certainty are depicted in \ref{im: figure3}.

\subsection{Diagnostics model pipeline (SeMaCNN model)}
\begin{figure}[ht]
\center{\includegraphics[scale=1]{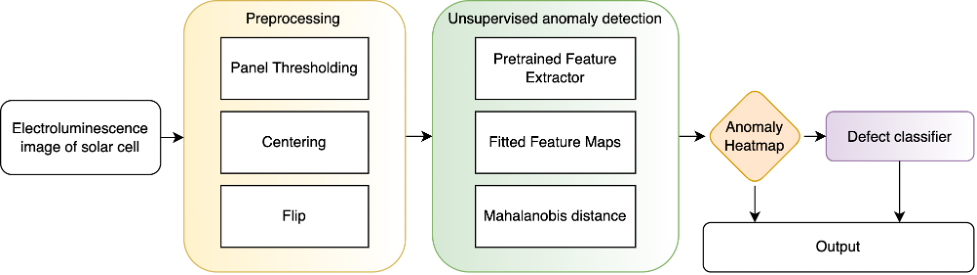}}
\caption{Flowchart of the proposed pipeline. Electroluminescent images go through four main blocks: Preprocessing (dataset specific), unsupervised anomaly detection, generation of anomaly heatmaps, and classification.}
\label{im: figure4}
\end{figure}

The defect diagnostic model pipeline includes four sequential steps (\ref{im: figure4}): pre-processing, unsupervised anomaly detection, anomaly segmentation (heatmap generation) and defect classification. Pre-processing steps of EL images differ for HEPV and ELPV datasets since datasets were obtained under different conditions. Pre-processing consists of geometric image transformations like thresholding, centering to select only an area with a cell on raw image and flipping for data augmentation. 
The unsupervised anomaly detection algorithm based on PaDiM\cite{defard_padim_2020} produces an anomaly heatmap. This heatmap can be used for semantic segmentation or passed to the defects classifier. The classifier consists of a neural network trained via supervised learning on manually annotated subset of images that separates generated heatmaps into classes. Defect classification via unsupervised anomaly detection is the general advantage of the proposed model. In contrast to supervised models that require the whole library of annotated images with various kind of defects, our approach is concentrate on detection of areas in EL images with anomalous (unusual) patterns.

\subsubsection{Pre-processing}
The preprocessing pipeline has been designed separately for each dataset.

\paragraph{HEPV Dataset}
The HEPV dataset contains raw EV images with 8-bit brightness channel and 720×720 px resolution and their pre-processing includes normalization, edges removal and data augmentation. Image borders have been detected by thresholding pixel intensity values at level below 50 and removed (see Figure S1 in Supplementary materials). Non-orthogonal of skewed border lines were eliminated with the Douglas-Peucker algorithm to extract a region of interest. To augment data, semi-orthogonal transformations have been applied during the model fitting such as horizontal and vertical flips with a probability of 0.5 and rotation around the center point with a random angle.

\paragraph{ELPV Dataset}
Images in the ELPV dataset were already pre-processed and no centering or area selection were required. Due to the limited number of images in the ELPV dataset, we have created additional data via data augmentation techniques: rotation of the image for an angle up to 90 degrees and random flip along both the horizontal and vertical axes. Up to 10 transformations have been applied to each image which correspondingly increased the dataset size.

\subsubsection{Anomaly detection and classification by semi-orthogonal embedding for PaDiM}
\begin{figure}[ht]
\center{\includegraphics[scale=1]{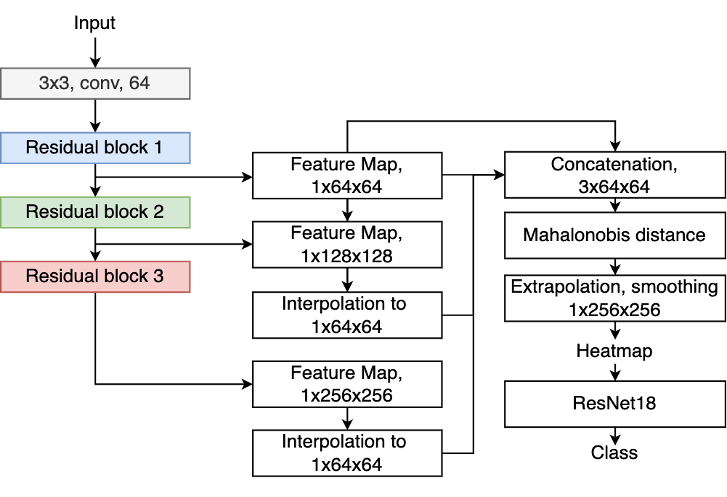}}
\caption{Anomaly detection block architecture. Feature maps are extracted after each residual block of ResNet18. All feature maps are interpolated and concatenated into one vector. Mahalanobis distances are calculated for each image pixel and then converted to anomaly heatmaps. Heatmaps in turn are classified by pre-trained ResNet18}
\label{im: figure5}
\end{figure}

For the anomaly segmentation, we have used the PaDiM model\cite{defard_padim_2020} extended with semi-orthogonal embedding\cite{kim_semi-orthogonal_2021}. PaDiM takes into account correlations among semantic levels of a pretrained CNN\cite{defard_padim_2020}. The ResNet18 model pre-trained on ImageNet was used as feature extractor\cite{deng_imagenet_2009} for generating relevant features (\ref{im: figure5}). A set of images without defects has been selected as a training subset. Training procedure consisted of creating a normal distribution of a dataset. After applying augmentation and pre-processing described above, images have been put through a feature extractor. Feature maps have been extracted after each residual block and interpolated to the same size 64×64 px. For feature selection from the list of feature maps, the random feature selection method with semi-orthogonal embedding for approximation has been used as presented in Kim et. al.\cite{kim_semi-orthogonal_2021} Hence, the anomaly represents itself as the value outside of the Gaussian distribution of the dataset. 
Semi-orthogonal embedding generalizes the idea of the random feature selection as a low-rank approximation of a precision matrix for the Mahalanobis distance\cite{vareldzhan_anomaly_2021,yang_visual_2021}. Mahalanobis distance is a metric that measures standard deviations between a given sample and a distribution of normal samples. Mahalanobis distance is supposed to have unimodal Gaussian distributions if the spatial alignment of samples is done by pre-processing. Thereby the training process consisted of computation of the normal distribution of the dataset.
To evaluate each patch of the image during the inference, we applied the Mahalanobis distance method. It represents the distance between patch embedding created by a feature extractor and the normal distribution that has been calculated during the training phase. The matrix of distances forms an anomaly map with high values corresponding to abnormal areas. 
For the classification of anomaly heatmaps, we have adopted the ResNet18 model, a shallow deep learning model widely used for computer vision tasks.

\subsection{Alternative methods}
To benchmark the proposed model, we compare it with other unsupervised and supervised methods.
\paragraph{Unsupervised methods}
We compare the SeMaCNN model with a set of widely used unsupervised anomaly detection models. One of the more basic ones is the traditional Fully Convolutional Autoencoder as presented elsewhere\cite{ji_invariant_2019,lai_robust_2019}. After creating an anomaly heatmap, an amplitude binary and/or Otsu thresholds\cite{otsu_threshold_1979} have been used. Experiments with Single Value Decomposition as well as K-means resulted in low accuracy and F1.
\paragraph{Supervised methods}
We have used supervised classification networks VGG-11[27]\cite{simonyan_very_2015}, ResNet50\cite{he_deep_2016}, ResNext101\cite{xie_aggregated_2017} to test the balance between the network depth (number of parameters in the neural network) and its performance on the HEPV dataset.

\section{Results}
\subsection{HEPV Dataset}
A set of models has been trained on A/BC sub-splits of the primary dataset. Best result for each optimized model is given in the \ref{table: models}. SeMaCNN demonstrated best metrics among all tested models reaching accuracy 0.852 and F1 0.888 respectively (Table \ref{table: models}). 

In order to compare energy efficiency of models we accounted energy consumption and carbon emissions during training of the model (applying regional specific carbon intensity coefficient equal to 240.56~g/kWh) with open-source python library \textit{eco2AI}\cite{eco2ai_lib}. As one can see in Table \ref{table: models} carbon emissions of SeMaCNN equals to 92.9 g that is much lower than for ResNet-50.

\begin{table}[ht]
    \centering
    \caption{Best experimental results for composite SeMaCNN, semi-orthogonal, convolutional and autoencoder models trained on A/BC dataset}
    \begin{center}
    \begin{tabular}{c p{3cm} c c c}
    \hline
        \textbf{№} & \textbf{Model} & \textbf{Accuracy} & \textbf{F1} & \textbf{$CO_2$ emissions, g} 
        \\ \hline
        1 & \textbf{SeMaCNN} & \textbf{0.852} & \textbf{0.888} & 92.9
        \\ 
        2 &	Semi-orthogonal, thresholding &	0.721 &	0.709 & 25 
        \\
        3 &	ResNet-50 &	0.808 &	0.821 & 169.9
        \\
        4 &	Autoencoder &	0.693 &	- & 83.6
        \\ \hline
    \end{tabular}
    \end{center}
    \label{table: models}
\end{table}

\begin{figure}[ht]
\center{\includegraphics[scale=1]{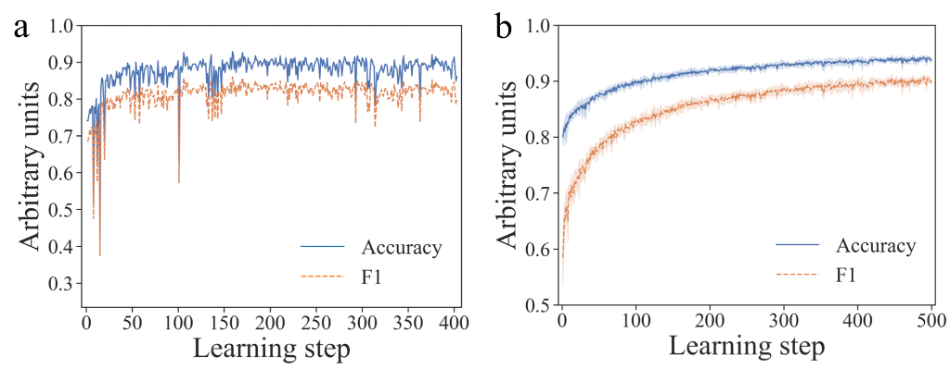}}
\caption{Evolution of the accuracy and F1 metrics during training of the classifier on the heatmaps with respect to the number of steps for: a) HEPV dataset, b) ELPV dataset.}
\label{im: figure6}
\end{figure}

Images of type B contain only slight deterioration spots, therefore separation of images with such defects is quite ambiguous. SeMaCNN trained on A/C subset on train/test split for 400 steps (\ref{im: figure6}a) reached much higher maximum accuracy and F1 equal to 0.951 and 0.961 respectively.

\begin{table}[ht]
\caption{Comparison of SeMaCNN performance for various training subsets, gray boxes indicate type subset selection for binary classification.}
\center{\includegraphics[scale=0.5]{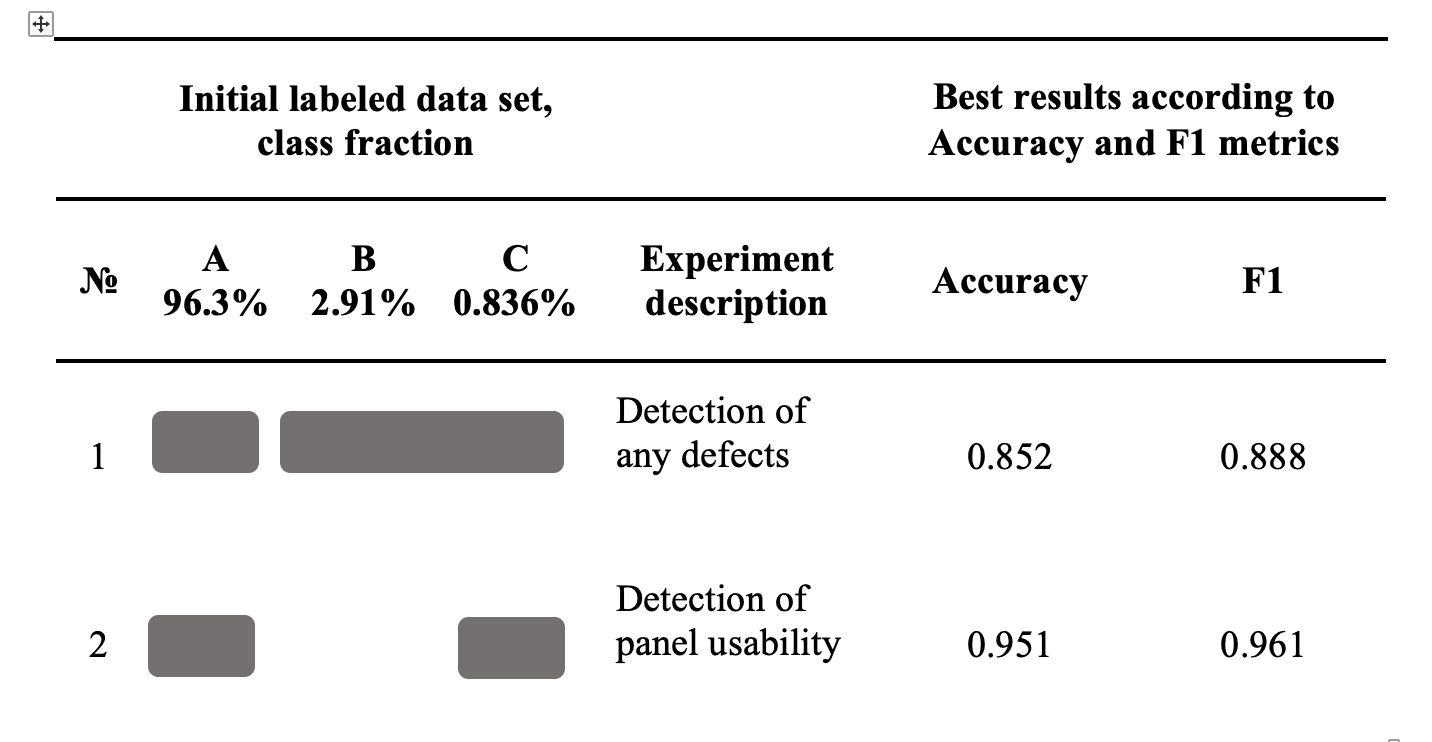}}
\label{im: table2}
\end{table}

\begin{figure}[ht]
\center{\includegraphics[scale=1]{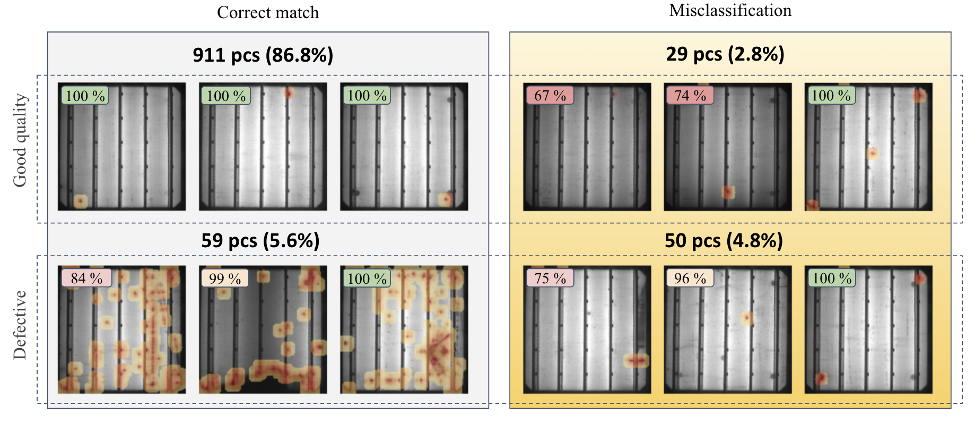}}
\caption{Qualitative results of a prediction made by the semi-orthogonal+ResNet18 model for good quality and defective cells arranged according to confusion matrix: Correct match (left panel and gray background), Misclassification (right panel, and yellow background). Values above each row indicate the number (share) of images of the corresponding class, values within cells indicate prediction probability of correct classification by the model, anomalies associated with defects highlighted in images by the yellow-red colormap areas.}
\label{im: figure7}
\end{figure}

The SeMaCNN distinguishes scratches and localized dark areas, spots and scratches with high fidelity as presented in Figure 7. At the same time, SeMaCNN experiences difficulties in images where black areas are taking up a big part of the screen or panels are dimmed, however these images represent only a minor fraction of the whole the dataset. As presented in \ref{im: figure7}, the percentage of correctly identified images is still high reaching 92.4\%.

\subsection{HEPV Dataset}
To compare results with currently existing models, we have applied the SeMaCNN to a public ELPV dataset. The proposed semi-supervised model has managed to produce results comparable to current SOTA (see Table S3 in Supplementary Materials) as presented in Demirchi et. al.\cite{demirci_efficient_2021} where authors used a 2000 feature subset out of 5120 feature dimensions. These dimensions are produced using four different feature extraction networks (DarkNet-19\cite{redmon_yolo9000_2017}, VGG-16\cite{simonyan_very_2015}, VGG-19\cite{simonyan_very_2015}, ResNet50\cite{he_deep_2016}). Our solution uses ResNet18\cite{he_deep_2016} as a backbone that provides comparable accuracy with fewer operations. Both our average and worst results achieved during the 4-fold cross-validation (Table S2) are comparable, which confirm the quality of our model (\ref{im: figure6}b). The proposed model has also shown great performance on the EPFL dataset\cite{buerhop-lutz_benchmark_2018} and accuracy and F1 equal 94.6\% and 91.1\%, respectively. Thus, SeMaCNN demonstrated good balance between complexity and performance across all published models for this dataset (Table S3) demonstrating best accuracy over all models except having twice as many parameters (Table S4).

\section{Conclusions}

In this study, we proposed a novel semi-supervised model SeMaCNN for automatic defects detection of heterojunction solar cells by processing EL images. The model extracts semi-orthogonal embedding of Mahalanobis distances via unsupervised learning and then passes it to trained supervised convolutional neural networks. The model solves both defect segmentation and cell quality classification tasks. 
The model has been trained on the in-plant data from 68 748 samples of monocrystalline solar cells and achieved accuracy and F1 score equal to 95.8\% and 92.5\%, respectively for the task of binary classification of solar cells quality, respectively. To validate the model and to define the extent of industrial potential, we also tested it on the public dataset ELPV and achieved comparable accuracy and F1 score equal 94.6\% and 91.1\%, respectively.
The SeMaCNN model demonstrates a good balance between its performance and computational costs: the model performance is superior or comparable to the widely used supervised CNN and unsupervised autoencoder models with less computational costs for features extraction. The achieved balance between performance and computational costs makes the SeMaCNN model applicable for integrating into quality control systems of solar cell manufacturing with further potential extension to defects causes revealing.

\section*{Acknowledgments}
The authors thank the Sber ESG Direction and Sber CIB (Key Clients department, Power and Utilities Coverage) for providing expert consulting in the domain of research.

\section*{Competing interests}
The authors declare no competing interests.

\section*{Data availability statement}
Electroluminescence Photovoltaic images Database was provided by LCC “Unigreen Energy”. Research data is not shared.

\bibliographystyle{unsrt}
\bibliography{main}

\end{document}